\documentclass[conference]{IEEEtran}
\IEEEoverridecommandlockouts
\usepackage{cite}
\usepackage{amsmath,amssymb,amsfonts}
\usepackage{algorithmic}
\usepackage{graphicx}
\usepackage{textcomp}
\usepackage{multirow}
\usepackage{xcolor}

\def\BibTeX{{\rm B\kern-.05em{\sc i\kern-.025em b}\kern-.08em
    T\kern-.1667em\lower.7ex\hbox{E}\kern-.125emX}}
\begin{document}

\title{Never Reset Again: A Mathematical Framework for Continual Inference in Recurrent Neural Networks

\thanks{The authors gratefully acknowledge the helpful discussions with Sander Bohte and Simeon Kanya. This work has been funded by the Dutch Organization for Scientific Research (NWO) with Grant KICH1.ST04.22.021}
}

\author{\IEEEauthorblockN{Bojian Yin}
\IEEEauthorblockA{\textit{Electrical Engineering} \\
\textit{Eindhoven University of Technology}\\
The Netherlands \\
b.yin@tue.nl}
\and
\IEEEauthorblockN{Federico Corradi}
\IEEEauthorblockA{\textit{Electrical Engineering} \\
\textit{Eindhoven University of Technology}\\
The Netherlands \\
f.corradi@tue.nl}
}

\maketitle
\vspace{-100pt}  

\begin{abstract}

Recurrent Neural Networks (RNNs) are widely used for sequential processing but face fundamental limitations with continual inference due to state saturation, requiring disruptive hidden state resets. However, reset-based methods impose synchronization requirements with input boundaries and increase computational costs at inference. To address this, we propose an adaptive loss function that eliminates the need for resets during inference while preserving high accuracy over extended sequences. By combining cross-entropy and Kullback-Leibler divergence, the loss dynamically modulates the gradient based on input informativeness, allowing the network to differentiate meaningful data from noise and maintain stable representations over time. t our reset-free approach  outperforms traditional reset-based methods when applied to a variety of RNNs, particularly in continual tasks, enhancing both the theoretical and practical capabilities of RNNs for streaming applications.

\end{abstract}

\begin{IEEEkeywords}
RNNs, state saturation, State Resetting, Loss Function
\end{IEEEkeywords}

\section{Introduction}
Recurrent neural networks (RNNs) constitute the foundation for processing sequential data in deep learning and are the natural choice for time series modelling~\cite{elman1990finding, hopfield1982neural}. 
Over time, these architectures have evolved into specialized variants designed to address a variety of computational tasks. Notably, Gated Recurrent Units (GRUs) \cite{cho2014learning} optimize gradient flow, Linear Recurrent Networks including State Space Models (SSMs) \cite{gu2021efficiently,orvieto2023resurrecting,smith2022simplified} enable efficient and parallel training of large datasets, and Spiking Neural Networks (SNNs) implement biologically plausible, energy-efficient computation for edge devices \cite{neftci2019surrogate,yin2021accurate}. 
Currently, RNN principles are applied to modern Large Language Models (LLMs), which, although based on transformer architectures \cite{vaswani2017attention}, act as RNNs during inference by processing tokens sequentially with cached states for computational efficiency \cite{liu2019roberta, brown2020language}.
Despite these architectural advancements in RNNs, fundamental challenges remain when continuously processing input data streams. In these scenarios, networks are required to generate stable representations and accurate predictions while processing uninterrupted data streams that substantially exceed typical training sequence lengths. This must be achieved without compromising inference performance.

A major challenge in continual inference settings is the state saturation problem, which significantly limits the practical deployment of RNNs \cite{merrill2021formal,chen2024stuffed}.
This challenge originates despite their theoretical Turing completeness, which highlights their potential to solve a diverse range of tasks~\cite{siegelmann1995computational, chung2021neural}. This phenomenon manifests as progressive accuracy degradation during prolonged input exposure, affecting all RNN variants: traditional architectures, SNNs, and even LLM models~\cite{merrill2021formal,chen2024stuffed}. The degradation is particularly apparent in continuous speech recognition and streaming tasks \cite{kalchbrenner2016neural, yindynamic}, where performance declines gradually with exposure to long input speech sequences. 

The mechanism that leads to state saturation is the temporal accumulation of information within the hidden states. During continual operation, this accumulation generates interference between historical and present information, progressively degrading the network's capacity to process new inputs \cite{tay2020long,merrill2021formal,chen2024stuffed}. Recent theoretical advances have provided insights into this limitation. Merrill \cite{merrill2021formal} used language theory to formally analyze the behavior of saturated networks, while Panahi et al. \cite{panahi2021generative} explained the implicit additive properties of RNNs by incorporating biologically-inspired saturation bounds. The significance is further highlighted by Paassen et al.'s \cite{paassen2021reservoir} that have demonstrated that RNNs can emulate any finite state machine by mapping neurons to states.
Current methods to address state saturation involve state collapse prevention via continual pre-training on longer sequences \cite{chen2024stuffed} and dynamic state resets using an extra action selection circuit \cite{yindynamic}. Other methods use periodic or dynamic hidden state resets but struggle with the need for synchronization with input boundaries, often missing in real-world scenarios \cite{havard2020catplayinginthesnow}. While distinct sample boundaries are manageable in controlled settings, they're typically absent in practical streaming and edge deployments \cite{gu2022efficient,yindynamic}. Despite these efforts, state saturation still challenges RNNs during long inferences, impacting their ability to process continuous data streams reliably.

These observations motivate our central research question:
\begin{quote}
\textit{``Can we develop a method for training RNNs that eliminates the need for resetting during continual inference?''}
\end{quote}

This study offers a mathematical approach to the state saturation problem, eliminating the need for hidden state resets. We present an adaptive loss function that adjusts learning dynamically, using categorical cross-entropy to assess the target patterns and Kullback-Leibler (KL) divergence \cite{cui2023decoupled} to build a uniform distribution for noisy or irrelevant input. This approach enables autonomous output probability adjustment while preserving hidden state continuity.


The main contributions of this work are:
\begin{itemize}
    \item 
    a formal characterization of how prolonged input sequence affects RNNs dynamics in continual inference, including vanilla RNNs, GRU, SSMs, SNNs.
    \item 
    a mathematical analysis demonstrating saturation prevention and gradient flow maintenance in the absence of hidden state resets.
    
    \item 
    the development of a loss function that combines cross-entropy and KL divergence terms and enables stable and accurate continual inference without state reset.
\end{itemize}


\begin{figure*}[t]
	\centering
	\includegraphics[width=1.\textwidth]{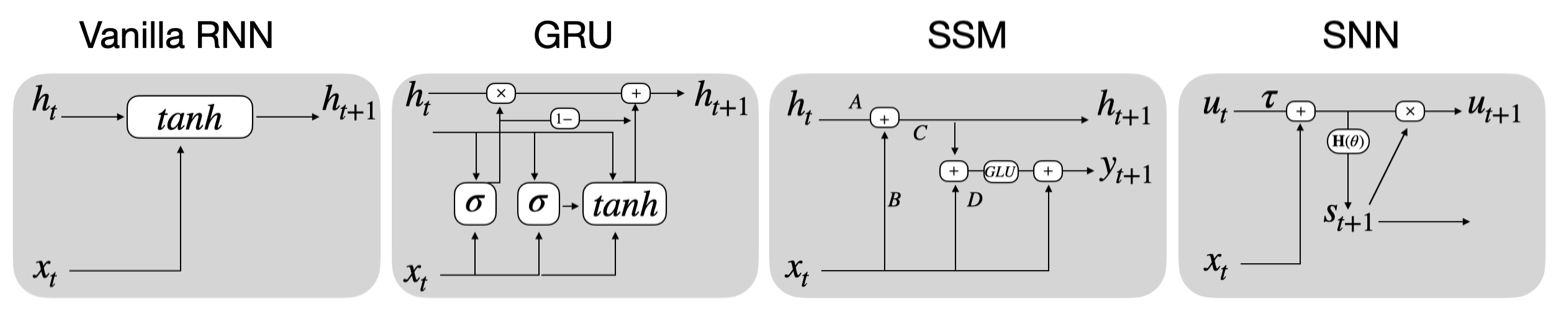}
	\caption{Computational graphs of different recurrent architectures.}
	\label{fig:rnns}
\end{figure*}

\section{Background}\label{background}

The sequential modelling performance of recurrent architectures is fundamentally constrained by and also benefits from their hidden dynamics  \cite{paassen2021reservoir,merrill2021formal,chen2024stuffed}. This section examines the theoretical foundations and use of state reset mechanisms, contextualizing their role in mitigating saturation phenomena.

Sequential data processing in deep learning is described by the general recurrent framework $\mathbf{h}_t = f(\mathbf{h}_{t-1}, \mathbf{x}_t)$, where $\mathbf{h}_t$ represents the hidden state at time $t$, and $f(\cdot)$ denotes a learnable state transition function operating on the previous state $\mathbf{h}_{t-1}$ and current input $\mathbf{x}_t$. This framework is commonly used in RNNs and for applications in many computational tasks. For example, vanilla RNNs implement this framework through $\mathbf{h}_t = \tanh(\mathbf{W}_h\mathbf{x}_t + \mathbf{U}_h\mathbf{h}_{t-1} + \mathbf{b}_h)$, where $\{\mathbf{W}_h, \mathbf{U}_h, \mathbf{b}_h\}$ constitute the learnable parameters. To solve the vanishing gradient problem inherent in vanilla RNNs, GRU units have been introduced with adaptive gating mechanisms: $\mathbf{r}_t = \sigma(\mathbf{W}_r\mathbf{x}_t + \mathbf{U}_r\mathbf{h}_{t-1} + \mathbf{b}_r)$ and $\mathbf{z}_t = \sigma(\mathbf{W}_z\mathbf{x}_t + \mathbf{U}_z\mathbf{h}_{t-1} + \mathbf{b}_z)$, facilitating controlled information flow through $\mathbf{h}_t = (1-\mathbf{z}_t)\odot\mathbf{h}_{t-1} + \mathbf{z}_t\odot\tanh(\mathbf{W}_h\mathbf{x}_t + \mathbf{U}_h(\mathbf{r}_t\odot\mathbf{h}_{t-1}) + \mathbf{b}_h)$. Additionally, SSMs introduce a continuous-time perspective because they formulate the time dynamics as $\dot{\mathbf{h}}(t) = \mathbf{A}\mathbf{h}(t) + \mathbf{B}\mathbf{x}(t)$, $\mathbf{y}(t) = \mathbf{C}\mathbf{h}(t) + \mathbf{D}\mathbf{x}(t)$, where $\{\mathbf{A}, \mathbf{B}, \mathbf{C}, \mathbf{D}\}$ represent learnable state-space parameters for long-range dependency. In addition, SNNs introduce bio-inspired principles and temporal processing by relying on the neuron's membrane potential dynamics $\mathbf{u}_t = \alpha\mathbf{u}_{t-1} + \sum_i w_i\mathbf{s}_{t-1}^i$, coupled with a threshold-based spike generation mechanism $\mathbf{s}_t = H(\mathbf{u}_t - \theta)$, where $H$ denotes the Heaviside function and $\theta$ represents the firing threshold. Each architectural variant in this unified framework offers unique characteristics in modeling capacity, biological plausibility, and computational efficiency, allowing for tailored solutions in sequential data processing.

\subsection{Formalization of Reset Mechanisms}

In general, Reset techniques in RNNs can be formalized by state transformation operations. The reset function $\Phi: \mathbb{R}^d \rightarrow \mathbb{R}^d$ operates on hidden states $h_t \in \mathbb{R}^d$ according to:
\begin{equation}
    h_t = \Phi(h_{t-1}, r_t) = r_t \odot h_{t-1}
    \label{eq:reset}
\end{equation}
where $r_t \in \{0, 1\}^d$ denotes a binary reset mask, and $\odot$ denotes the Hadamard product. This formulation requires a priori knowledge of sequence boundaries---an assumption that proves problematic in continuous processing paradigms \cite{pascanu2013difficulty, bengio1994learning}.

This reset operation breaks the recurrent dynamic by enforcing $\mathbf{h}_t \leftarrow \mathbf{h}_0$, 
where $\mathbf{h}_0$ typically follows either zero initialization ($\mathbf{h}_0 = \mathbf{0}$) or random initialization ($\mathbf{h}_0 \sim \mathcal{N}(0, \sigma^2\mathbf{I})$). The Default reset method in this paper is random initialization.

\subsection{Gated Architectures and Soft Reset Functions}

Some recurrent neural network architectures incorporate learnable reset mechanisms through gated structures \cite{cho2014learning, gers2000learning}. For example, the GRU \cite{cho2014learning} introduces a reset gate formulation as follows:
\begin{equation}
    r_t = \sigma(W_r[h_{t-1}; x_t] + b_r)
    \label{eq:gru}
\end{equation}
where $W_r \in \mathbb{R}^{d\times(d+n)}$ and $b_r \in \mathbb{R}^d$ are the learnable parameters, and $[\cdot;\cdot]$ denotes vector concatenation. The same idea extends to LSTMs~\cite{gers2000learning} that exploit a forget gate mechanism with context information, which learns when to reset the network state during continual input processing, learning longer temporal dependencies from the data.

Recent methods introduce surprisal-driven feedback~\cite{rocki2016surprisal} and binary input gated~\cite{li2019reading} mechanisms in recurrent networks to enable effective resetting of predictions and selective processing during interference. However, these solutions address a different problem because they focus on learning more complex temporal patterns and long-term dependencies, rather than tackling the issue of state saturation during inference in continuous operations \cite{hochreiter1997long}.

\subsection{Theoretical Framework}\label{sec:theory}

The theoretical foundations of reset operations can be examined using dynamical systems theory. Current theories focus more on discrete-time scenarios, leaving continuous operational dynamics an area of ongoing exploration \cite{siegelmann1995computational, chung2021neural, chu2024lyapunov}. Let $\mathcal{H}$ denote the hidden state manifold, where the reset operation induces a transformation $T: \mathcal{H} \rightarrow \mathcal{H}$. To establish the mathematical necessity of hidden state resets in RNNs, where $T(h_t)=h_0$, we examine two within this dynamical systems~\footnote{More details in appendix}:

\begin{enumerate}
    \item \textbf{Lyapunov Stability:} A fixed point $h^* \in \mathcal{H}$ exhibits Lyapunov stability \cite{chu2024lyapunov} if, for every $\varepsilon > 0$, there exists $\delta > 0$ such that:
    \begin{equation}
        \|h_0 - h^*\| < \delta \implies \|h_t - h^*\| < \varepsilon, \quad \forall t \geq 0.
    \end{equation}
    The reset operation to $h_0$ maintains the system state within a stable neighborhood of fixed point $h^*$, eliminating the accumulation of computational artifacts that could misleading future processing.

    \item \textbf{Information Preservation:} Given the mutual information $I(x_t; h_{t-1})$ between input $x_t$ and the previous hidden state $h_{t-1}$, we observe that over extended sequences:
    \begin{equation}
        \lim_{t \to \infty} I(\mathbf{x}_t; \mathbf{h}_t|\mathbf{X}_{0:t-1}) \to 0.
    \end{equation}
    Consequently, reset operations must satisfy an information preservation criterion:
    \begin{equation}
        I(x_t; T(h_t)) \geq \gamma I(x_t; h_{t-1}), \quad \gamma \in (0,1],
    \end{equation}
    This inequality holds because resetting terminates dependencies on $h_{t-1},h_{t-2},\dots $, 
    ensuring the retention of essential temporal dependencies while facilitating effective processing of current inputs $x_t$ through hidden state re-centering.
\end{enumerate}







The current reset method $\mathbf{h}_t \leftarrow \mathbf{h}_0$ has limitations limiting RNNs' practical use. A key challenge is the need for explicit alignment with input boundaries. Furthermore, the implementation of dynamic reset mechanisms introduces non-negligible computational overhead \cite{kalchbrenner2016neural,yindynamic}. 

\section{Theory of Reset-free Loss}\label{methods}

In this section, we present the detailed formulation of the proposed loss function and explain how it addresses the state saturation problem in RNNs during continual inference without the need for resetting the hidden states.


Consider a sequence of input data $\{\mathbf{x}_t\}_{t=1}^T$ and corresponding target labels $\{y_t\}_{t=1}^T$, where $T$ is the sequence length. The RNN processes the input sequence to update hidden states $\{\mathbf{h}_t\}_{t=1}^T$ and outputs probability distributions over $C$ classes at each time step $t$:
\begin{equation}
\mathbf{p}_t = [p_{t,1}, p_{t,2}, \dots, p_{t,C}]^\top = \text{softmax}(\mathbf{W}_o \mathbf{h}_t + \mathbf{b}_o),
\end{equation}

where $\mathbf{W}_o$ and $\mathbf{b}_o$ are the output weight matrix and bias vector, respectively.

Resetting hidden states in traditional RNNs is done before new input sequences.

To address this, hidden states are reset (see Section \ref{sec:theory}) to enhance relevant information and ensure $p_t$ focuses on the current input $x_t$ with a random state $h_0$. Thus, to avoid explicit state reset, we need to formulate a loss function that enables the model to learn from data and adjust output probabilities amidst noise \textbf{without resetting the hidden states}, ensuring temporal coherence. Our objectives are:

\begin{itemize}
    \item \textbf{Accurately predict the true class $y_t$ when $m_t = 1$.}
    \item \textbf{Output a uniform distribution over classes when $m_t = 0$,} indicating maximum uncertainty, effectively resetting the output probabilities without reinitialize the hidden states.
\end{itemize}

We firstly introduce a mask $\{m_t\}_{t=1}^T$ that indicates whether each time step $t$ corresponds to informative data ($m_t = 1$) or noise ($m_t = 0$). Then, we define the total loss $L_{\text{total}}$ with two parts:

\begin{enumerate}
    \item \textbf{Categorical Cross-Entropy Loss} for informative inputs ($m_t = 1$):
    \begin{equation}
    L_{\text{CE}}(t) = -\sum_{k=1}^C y_{t,k}\log p_{t,k}
    \end{equation}

    \item \textbf{Kullback-Leibler Divergence Loss} towards a uniform distribution for noise inputs ($m_t = 0$):
    \begin{equation}
    L_{\text{KL}}(t) = D_{\text{KL}}(\mathbf{p}_t \parallel \mathbf{u}) = \sum_{k=1}^C p_{t,k} \log \left( \frac{p_{t,k}}{v_k} \right),
    \end{equation}

    where $\mathbf{u}$ is the uniform distribution over classes, i.e., $v_k = \dfrac{1}{C}$ for all $k$.
\end{enumerate}

Finally, the total loss over the sequence is given by:

\begin{equation}
L_{\text{total}} = \sum_{t=1}^T \left[  m_t \cdot L_{\text{CE}}(t) + (1 - m_t) \cdot L_{\text{KL}}(t) \right].
\end{equation}

Our proposed objective function adaptively initializes states in recurrent architectures using a dual optimization. It jointly minimizes the cross-entropy while constraining the hidden state distribution via Kullback-Leibler divergence regularization. This approach lets the network adjust its internal states automatically, preventing state saturation without needing explicit resets.
Moreover, this ensures the model retains hidden states to smoothly transition from noisy input to new data, avoiding saturation and ensuring continuous gradient flow.

\section{Experiments of continuous streaming tasks}\label{experiments}

We evaluate the performance of our proposed loss function with sequential tasks that exceed the training length distribution, mimicking real-world applications in which the inference process cannot assume input data with clear separations as, for example, in audio applications (e.g., keyword spotting).  The experimental protocol involves testing the trained network on concatenated input sequences, where multiple training samples are combined to form an extended temporal sequence.

\subsection{Dataset}
\begin{figure}
	\centering
	\includegraphics[width=0.8\linewidth]{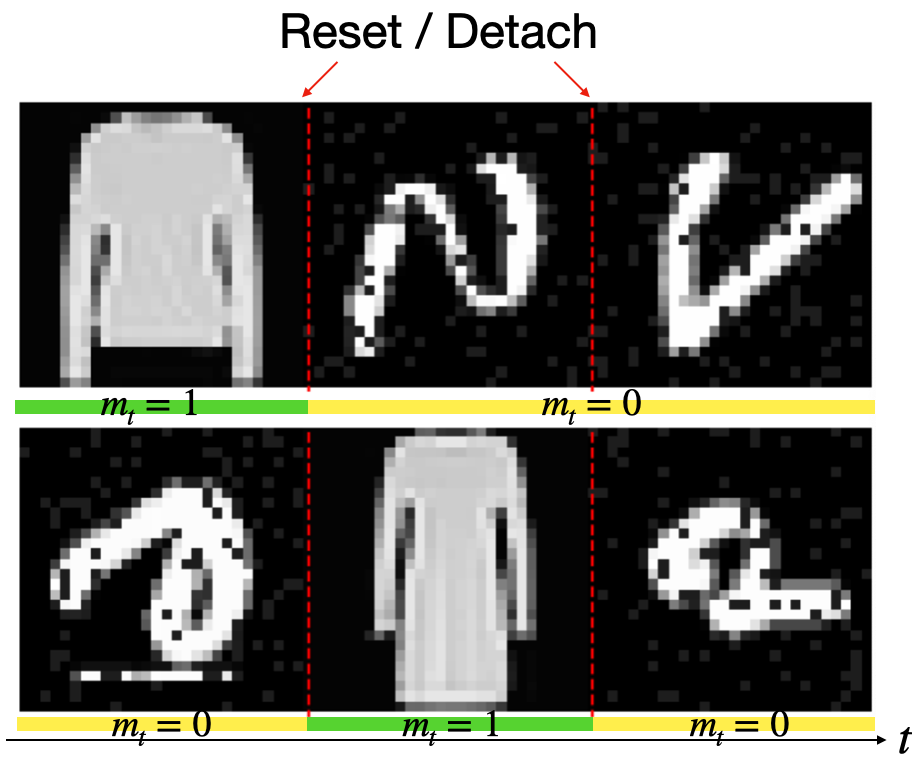}
	\caption{Example of \textit{Sequential FashionMNIST} with corresponding mask.}
	\label{fig:sfashion}
\end{figure}

We first examine our loss function on a cross-domain sequential learning task (\textit{Sequential Fashion-MNIST}) as illustrated in Fig.~\ref{fig:sfashion}. The experimental setup comprises 84-timestep sequences, where each sequence contains one Fashion-MNIST sample randomly interspersed with two MNIST digit sequences (p=0.1 dropout). Both datasets contain 60,000 training and 10,000 test instances across 10 categorical classes. Each 28×28 grayscale image is transformed into a temporal sequence with 28 time steps. During training, the network implements state reset or detach mechanisms at 28-timestep intervals for various loss functions. At inference, we assess continual inference on concatenated sequences without intermediate resets or detach, enabling quantitative measurement of cross-domain discriminative capability and temporal stability on expended sequences.

To allow selective learning of Fashion-MNIST patterns, we implement a binary masking mechanism across the 84-timestep sequences. The mask tensor ($m_t \in \{0,1\}, t\in\{1,\dots,84\}$) marks Fashion-MNIST segments with ones and MNIST segments with zeros, enabling targeted gradient flow, in Figure \ref{fig:sfashion}. For each sequence, we generate six possible permutations of segment arrangements, randomly selected per batch. The Fashion-MNIST target sequence maintains its temporal coherence while MNIST segments undergo random permutation and dropout (p=0.1) to prevent memorization. This masking strategy ensures the network learns to identify fashion-specific temporal patterns while treating MNIST segments as structured noise, effectively creating a controlled environment for evaluating temporal pattern recognition.

Our second benchmark utilizes the \textit{Google Speech Commands} dataset--GSCv2 \cite{warden2018speech}.  
The expanded GSCv2 comprises 35 keyword classes plus an "unknown" category, with 36,923 training and 11,005 test instances. The acoustic signals undergo spectral decomposition via Mel-frequency cepstral coefficient (MFCC) analysis, implemented through a bank of 40 second-order bandpass filters logarithmically spaced along the Mel-scale (20Hz-4kHz). Following spectral normalization by standard deviation, each instance is encoded as a sequence of 101 timesteps, where each timestep is represented by a 40×3 feature matrix capturing the temporal evolution of spectral characteristics.

In GSCv2, we evaluate two distinct approaches to generate temporal masks. First, following \cite{yindynamic} , we implement a temporal-intensity mask (\textbf{TI}) that captures local dynamics through a combination of average magnitude and variability. Specifically, for input signal $x_t$, we compute the local average $\mu_t= |x_t+x_{t-1}|/2$ and variability $\sigma_t = |x_t-x_{t-1}|$, which are combined to form a mask signal $tvar_t = \tanh(4\sigma_t\mu_t)$. The final mask $m_t$ is obtained through exponential smoothing with time constant $\tau$ as $m_t =m_{t-1} + (1-\exp(-1/\tau))(tvar_t-m_{t-1})$ where $\tau=5$. Second, we propose an energy-based approach that leverages the spectral characteristics of the MFCC features. The mask is derived from the log-scaled frame energy $e_t = \log(1 + \sum_{h} x_{t,h}^2)$, where $x_{t,h}$ represents the $h$-th MFCC coefficient at time $t$. Similar to the temporal-intensity approach, we apply exponential smoothing to obtain the final mask $m_t = m_{t-1} + (1-\alpha)(e_t-m_{t-1})$ with smoothing factor $\alpha=0.7$.

\begin{figure}[!t]
	\centering
	\includegraphics[width=0.6\linewidth]{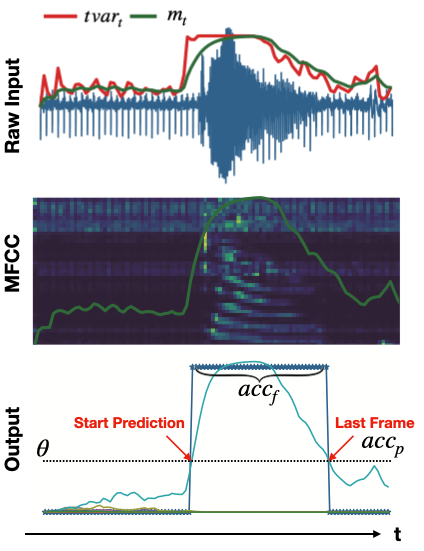}
	\caption{Online audio processing framework with masking and accuracy metrics. Top: Raw waveform with temporal variance ($tvar_t$, red) and its smoothed variant ($m_t$, green) serving as the mask signal, where $m_t > \theta$ defines active processing intervals. Middle: MFCC spectrogram with temporal mask application (green), illustrating feature extraction during active windows. Bottom: Decision metrics showing frame-wise accuracy ($acc_f$) across the processing duration and prediction accuracy ($acc_p$) at the final frame, bounded by activation threshold $\theta$ ($\theta=0.9$ in default). This demonstrates the temporal evolution of prediction confidence during continuous inference.}
	\label{fig:metrics}
\end{figure}
\subsection{Metrics}

To validate the performance, we evaluate across multiple recurrent architectures with our reset-free mechanism, including vanilla RNN, GRU, SSM, and SNNs, comparing their performance against traditional periodic reset approaches. The evaluation includes both discrete sample classification and continual inference setting. Let $\mathbf{m}_t \in \{0,1\}$ denote the binary indicator of pattern presence at time step $t$, and $\hat{y}_t, y_t$ represent the predicted and true labels, respectively. For single sample assessment, we quantified last frame prediction accuracy ($acc_p$) at the terminal point of feature presentation (illustrated in Fig.~\ref{fig:metrics}) as:
\begin{equation}
    acc_p = \frac{1}{N} \sum_{i=1}^N \mathbb{I}(\hat{y}_{t_i} = y_i) \cdot \mathbf{m}_{t_i}
\end{equation}
where $t_i$ corresponds to the \textit{last time step} where $\mathbf{m}_t = 1$ for the $i$-th sample, and $N$ is the total number of samples .

To measure the models' capacity for sustained operation, we measured the frame-wise accuracy ($acc_f$), computed as:
\begin{equation}
    acc_f = \frac{\sum_{t=1}^T \mathbb{I}(\hat{y}_t = y_t) \cdot \mathbf{m}_t}{\sum_{t=1}^T \mathbf{m}_t}
\end{equation}

where $\mathbb{I}(\cdot)$ is the indicator function. This dual evaluation function measures both the classification accuracy and the long-term temporal stability of the reset-free mechanism under different inference settings.

\section{Results}\label{results}
\begin{figure*}[h]
	\centering
    \includegraphics[width=1.\textwidth]{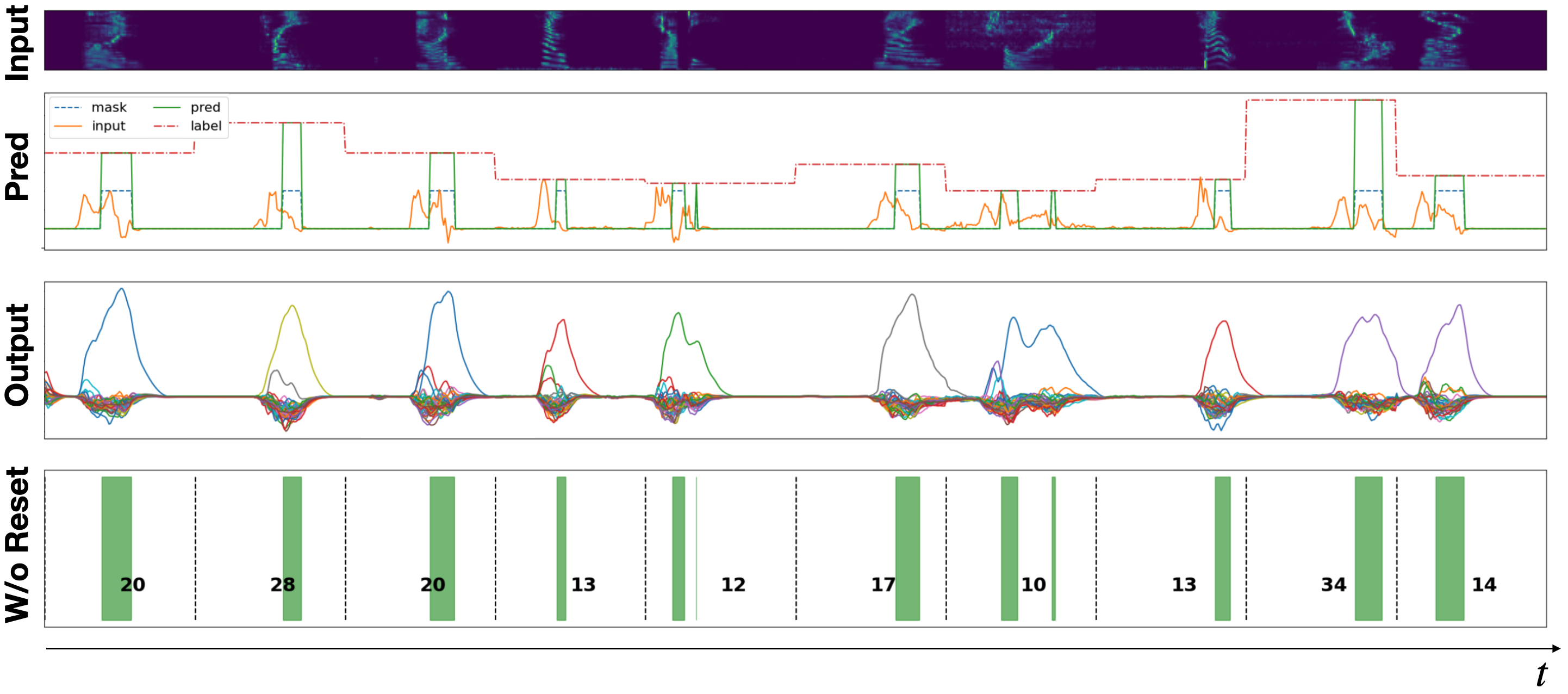}
	\caption{Visualization of reset-free GRU network dynamics on concatenated speech sequences trained with our proposed loss function. Top: MFCC spectrum of four concatenated speech utterances. Next is plotted the temporal mask as directly calculated and frame-wise network output and resulting classifications (green: correct label, red: incorrect label).}
	\label{fig:concat}
\end{figure*}

\subsection{Sequential FashionMNIST task}

Experimental results demonstrate the performance of our method in cross-domain sequence processing. We evaluate the proposed approach on the Sequential Fashion-MNIST task using a two-layer GRU with 256 hidden neurons at each layer, examining both last-frame ($acc_p$) and frame-wise ($acc_f$) classification performance across sequence lengths ranging from single samples to 128 concatenated sequences ($>10k$ timesteps). As shown in Table~\ref{tab:fashionmnist}, our method exhibits improved temporal stability compared to baseline approaches. The experimental results indicate that hidden state detachment significantly outperforms periodic reset by preserving temporal context while selectively regulating gradient flow. While periodic reset exhibits minimal accuracy degradation (-0.06\%), its last frame and frame-wise performance (74.47\%, 85.69\%) remain suboptimal due to complete state elimination in resetting. Under hidden state detachment, our newly proposed loss function achieves 88.43\% last-frame accuracy on single sequences and maintains 86.69\% accuracy with 128 concatenated sequences, yielding minimal performance degradation (-1.74\%). This contrasts with the masked cross-entropy (\textbf{mCE}), which exhibits significant decay (-3.72\%). The stability improvement becomes more pronounced in the reset configuration, where our method constrains accuracy degradation to -1.57\% versus -11.89\% for the mCE. The frame-wise accuracy metrics corroborate these findings, with our method consistently preserving temporal stability across extended sequences in this cross-domain learning paradigm.

\begin{table}[]
\centering
\caption{ Evaluation on Sequential Fashion-MNIST classification. Performance comparison across loss function variants during continual inference. Baseline  applied \textbf{Periodical reset} with unmarked cross-entropy on both training and inference,  \textbf{mCE}$=m_t\cdot L_{\text{CE}}(t)$, applies element-wise multiplication between cross-entropy and temporal mask, and \textbf{our} proposed loss denotes the Reset-Free approach in inference. \textbf{Diff} denotes the accuracy degradation between single-sequence evaluation and 128 concatenated sequences. Results are averaged over three independent runs.}
\begin{tabular}{|ll|l|l|l|l||l|}

\hline
\multicolumn{2}{|l|}{Seq length}                   & 1      & 2      & 8      & 128 & Diff\\ \hline
\multicolumn{7}{|l|}{Last Frame Acc $acc_p$}                   \\ \hline
\multicolumn{2}{|l|}{Periodical reset}             &  85.69 & 85.58  & 85.66  &  85.63  & -0.06 \\ \hline
\multicolumn{1}{|l|}{\multirow{2}{*}{\begin{tabular}[c]{@{}l@{}}train w/\\ Detach\end{tabular}}} & mCE & \textbf{88.48}  & 86.89  & 85.13  & 84.76 & -3.72  \\ \cline{2-7} 
\multicolumn{1}{|l|}{}                        & \textbf{our} &  88.43 & \textbf{87.89}  & \textbf{87.28}  & \textbf{86.69}  & -1.74 \\ \hline
\multicolumn{1}{|l|}{\multirow{2}{*}{\begin{tabular}[c]{@{}l@{}}train w/\\ Reset\end{tabular}}}  & mCE & 82.42  & 77.00  & 74.21  &  70.53 &  -11.89\\ \cline{2-7} 
\multicolumn{1}{|l|}{}                        & Our & 82.64  & 82.42  & 81.08   & 81.07 & -1.57  \\ \hline
\multicolumn{7}{|l|}{Framewise Acc $acc_f$}                   \\ \hline
\multicolumn{2}{|l|}{Periodical reset}              & 74.47  & 74.26  & 74.46  & 74.42 &  -0.05 \\ \hline
\multicolumn{1}{|l|}{\multirow{2}{*}{\begin{tabular}[c]{@{}l@{}}train w/\\ Detach\end{tabular}}} & mCE &  77.30 & 75.34  & 73.49  &  71.42& -5.88  \\ \cline{2-7} 
\multicolumn{1}{|l|}{}                        & \textbf{our} & \textbf{77.51}  & \textbf{77.21}  & \textbf{76.09}  &  \textbf{75.62} &  -1.89 \\ \hline
\multicolumn{1}{|l|}{\multirow{2}{*}{\begin{tabular}[c]{@{}l@{}}train w/\\ Reset\end{tabular}}}  & mCE & 69.94  & 65.43  & 62.43  &  60.99 & -8.95 \\ \cline{2-7} 
\multicolumn{1}{|l|}{}                        & Our & 70.78  & 69.39  & 68.51  &  68.72  & -1.06 \\ 
\hline
\end{tabular}

\label{tab:fashionmnist}
\end{table}

\subsection{Google Speech Command}
\paragraph{Single sample} The results on the GSCv2 dataset illustrate obvious performance disparities across different neural architectures and training methodologies (Table \ref{tab:gscv2_acc}). Our reset-free loss consistently demonstrates high performance across all architectures, with the GRU variant achieving the best accuracy of 87.61\% ($\pm0.009$). This represents a notable improvement over both the baseline cross-entropy (\textbf{CE}) (85.65\% $\pm0.013$) and mCE (87.04\% $\pm0.006$) approaches. The effectiveness of temporal-intensity masking is particularly evident in the RNN architecture, where the reset-free loss yields a substantial 16.81\% improvement over the baseline (from 66.44\% to 83.25\%). Similar performance enhancements are observed across SSM and SNN architectures, with consistent gains and minimal variance across multiple independent runs, substantiating the robustness of our proposed method.

\begin{table}[h]
    \centering
        \caption{Classification accuracy ($acc_p$) on GSCv2 dataset with temporal-intensity masking, averaged over three independent runs. All Networks are using same number of neurons. Baseline cross-entropy (\textbf{CE})$=L_{\text{CE}}(t)$, represents the standard cross entropy loss without masking, Standard deviations are shown in parentheses. }
    \begin{tabular}{|c|c|c|c|c|}
        \hline
        & RNN & GRU & SSM & SNN \\ \hline
        CE & 66.44$\pm$ 0.012 & 85.65$\pm$0.013 & 83.72$\pm$0.014 & 82.09$\pm$0.009 \\ \hline
        mCE & 82.75$\pm$0.006 & 87.04$\pm$0.006 & 83.94$\pm$0.006 & 82.33$\pm$0.017 \\ \hline
        our & 83.25$\pm$0.005 & 87.61$\pm$0.009 & 85.04$\pm$0.001 & 83.26$\pm$0.011 \\ \hline
    \end{tabular}

    \label{tab:gscv2_acc}
\end{table}

\paragraph{Continual inference} We evaluated our reset-free method in continual inference scenarios using experiments with sequence lengths from 2 samples (202 timesteps) to 128 samples (12,928 timesteps), as shown in Figure \ref{fig:concat}. The results demonstrate that our reset-free approach achieves performance parity with the periodical-reset method across all selected models, effectively overcoming the state saturation challenges inherent in sequential processing. The GRU architecture exhibits superior performance, maintaining approximately 87\% accuracy, although it displays increased sensitivity to training approaches, with conventional CE performance degrading to 35\% for extended sequences. Notably, the SSM demonstrates enhanced robustness, maintaining consistent performance across different loss functions, while RNN and SNN architectures exhibit analogous behavioral patterns, suggesting comparable temporal processing mechanisms. Traditional CE-loss shows systematic performance deterioration with increasing sequence length across all architectures, and while masked-loss offers modest improvements over the baseline, it fails to match the stability and performance metrics achieved by our proposed reset-free method. 
These findings highlight that the choice of loss functions plays a crucial role, comparable to the architectural selection, in maintaining performance integrity during extended sequence processing tasks.

\begin{figure*}[t]
	\centering
    \includegraphics[width=1.\textwidth,height=4cm]{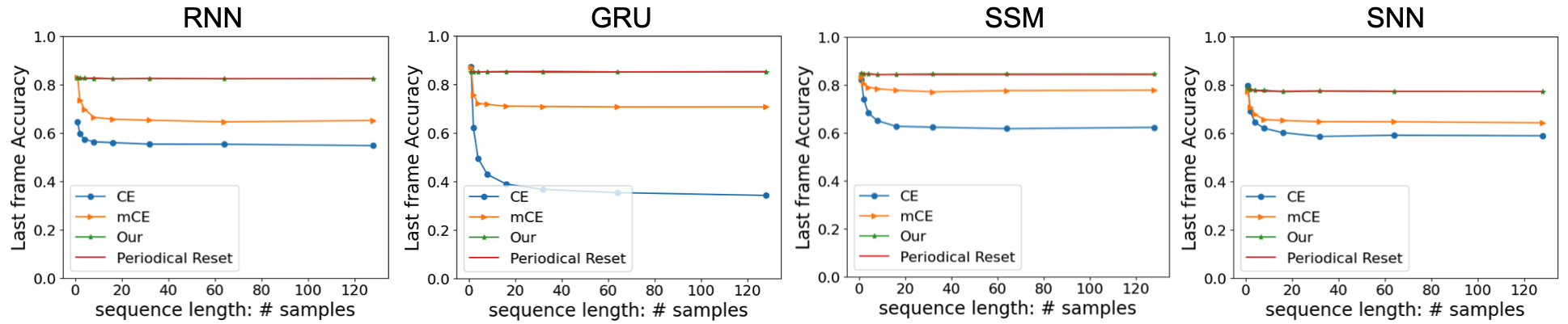}
	\caption{RNNs performance on GSCv2 across various neural architectures trained with different loss function and with variable sequence lengths. \textbf{Our} is overlapping with \textbf{Periodical Reset}.}
	\label{fig:concat}
\end{figure*}


\paragraph{Effects of different masking}

\begin{figure}[t]
	\centering
	\includegraphics[width=1.\linewidth]{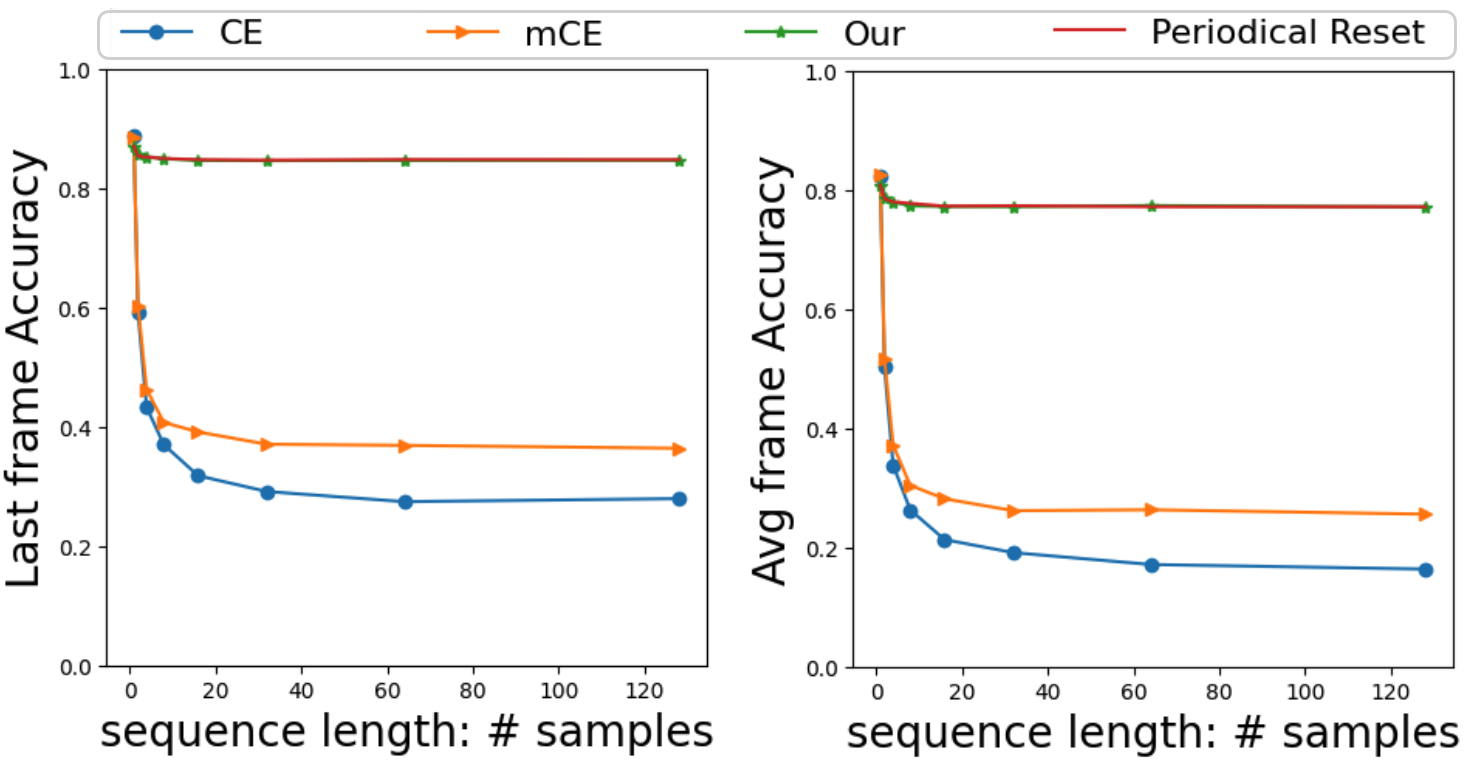}
	\caption{Performance comparison of different training methods using energy-based masking on extended sequences. Both metrics indicate that energy-based masking, while less robust than temporal-intensity masking, still enables our reset-free approach to maintain consistent performance without requiring periodic state resets.}
	\label{fig:energy_gsc}
\end{figure}

To investigate the impact of masking strategies on network performance, we perform  experiments comparing temporal-intensity and energy-based masking approaches on the GSCv2 dataset, as shown in Table~\ref{tab:energy_acc} and Figure~\ref{fig:energy_gsc}. Our analysis reveals that the choice of masking function significantly influences the network's learning dynamics and generalization abilities. While both masking strategies achieve comparable initial accuracy ($>87\%$) on single-sample sequences, their performance characteristics diverge substantially on extended sequences. The TI masking demonstrates superior robustness, particularly evident in our proposed reset-free method, which maintains consistent performance with minimal reduction (-0.42\%) across sequence lengths up to 128 samples, closely matching the periodical reset benchmark (-0.34\%) shown in Figure~\ref{fig:energy_gsc}. In contrast, energy-based masking exhibits substantial performance decrease (-2.49\% for reset-free, -2.86\% for periodical reset) on longer sequences. This difference becomes even more pronounced in conventional training approaches, where energy-based masking leads to substantial accuracy decay in both CE (-59.81\%) and mCE (-51.16\%) scenarios, compared to the relatively enhanced retention of performance with TI masking (-51.44\% and -16.3\%, respectively). These results highlight the role of masking function design in maintaining model performance across various sequence lengths, with TI masking emerging as the more effective approach for speech signal masking in the continual inference scenarios we have tested.

\begin{table}[]
\centering
\caption{Classification accuracy ($acc_p$) comparison between temporal-intensity (\textbf{TI}) and energy-based (\textbf{Energy}) masking strategies on GSCv2 dataset. Results are evaluated on concatenated sequences of varying lengths (1, 2, 8, and 128 samples) and averaged over three independent runs. \textbf{Reset} indicates periodical reset. \textbf{Diff} indicates the performance degradation from single to 128-sample sequences.}
\begin{tabular}{|ll||l|l|l|l|l|}
\hline
\multicolumn{2}{|l|}{}                              & 1 & 2 & 8 & 128 & Diff\\ \hline
\multicolumn{1}{|l|}{\multirow{2}{*}{CE}}  & TI     & 85.65  &62.10  &42.87  &34.21   & -51.44\\ \cline{2-7} 
\multicolumn{1}{|l|}{}                     & Energy & 87.74  & 59.13  &36.98  & 27.93 & -59.81\\ \hline\hline
\multicolumn{1}{|l|}{\multirow{2}{*}{mCE}} & TI     &87.04   &75.64  &71.79  &70.74   & -16.3\\ \cline{2-7} 
\multicolumn{1}{|l|}{}                     & Energy & 87.52  &60.16  &40.70  & 36.37  & -51.16\\ \hline\hline
\multicolumn{1}{|l|}{\multirow{2}{*}{Reset}}& TI    & 87.61  &87.19  &87.39  & 87.27   & -0.34\\ \cline{2-7} 
\multicolumn{1}{|l|}{}                     & Energy & 87.50  & 85.33 &85.14  &84.64  & -2.86\\ \hline\hline
\multicolumn{1}{|l|}{\multirow{2}{*}{Our}} & TI     & 87.61  &87.23  &87.27  &87.19   & -0.42\\ \cline{2-7} 
\multicolumn{1}{|l|}{}                     & Energy & 87.23  &85.64  &84.97  &84.74  & -2.49\\ \hline
\end{tabular}

\label{tab:energy_acc}
\end{table}

\section{Discussion}\label{discussion}

In this work, we demonstrated improvements in continuous sequence processing through three key contributions. Firstly, our reset-free methodology enables compact recurrent architectures to maintain stable performance during continual inference without performance decline. Secondly, the proposed dual-objective loss function achieves this by simultaneously optimizing for target data learning while enforcing divergence on non-relevant inputs, yielding performance metrics comparable to conventional reset-based approaches. Thirdly, through extensive empirical evaluation across diverse architectural paradigms—from classical RNNs to modern SSMs and biologically-inspired SNNs—we established the broad applicability of our method. The observed correlation between architectural complexity and temporal stability suggests fundamental principles governing the relationship between model complexity and continuous processing abilities. Notably, this methodology provides a robust foundation for temporal continuous learning by eliminating the necessity for timing resets in long sequence processing.

We observed that masking function design serves as one of the key factors of sequential learning performance. Our analysis reveals these functions act as guidance for learning, influencing the network's capacity to capture and preserve temporal dependencies. While this finding has substantial implications for architectural design in sequential learning systems, the current paradigm of manually engineered masking functions presents inherent limitations. This observation motivates the development of learnable masking mechanisms capable of automatic adaptation to underlying data distributions.

In a follow-up work, we plan to evaluate our approach in real-world continuous processing scenarios to validate its practical applicability, and possibly extending our methods to LLMs for addressing challenges in long-context inference and generation tasks. Additionally, the development of adaptive masking mechanisms, potentially incorporating transformer-style attention architectures or meta-learning frameworks, could lead to more efficient and interpretable solutions for continuous sequence processing. 

\bibliographystyle{IEEEtran}
\bibliography{references}

\clearpage 
\begin{appendices}
\section{Mathematical Analysis of State Saturation}

State saturation in RNNs can be formally analyzed through the lens of dynamical systems theory. Consider a general RNN with state evolution:

\begin{equation} \label{eq:general_rnn}
    \mathbf{h}_{t+1} = f(\mathbf{W}\mathbf{h}_t + \mathbf{U}\mathbf{x}_t + \mathbf{b})
\end{equation}

\noindent where $\mathbf{h}_t \in \mathbb{R}^n$ is the hidden state at time $t$, $\mathbf{x}_t \in \mathbb{R}^m$ is the input, $\mathbf{W} \in \mathbb{R}^{n \times n}$ is the recurrent weight matrix, $\mathbf{U} \in \mathbb{R}^{n \times m}$ is the input weight matrix, $\mathbf{b} \in \mathbb{R}^n$ is the bias term, and $f$ is an activation function.

\subsection{Linear RNNs}

For linear RNNs where $f$ is the identity function, the state evolution can be expressed as:

\begin{equation} \label{eq:linear_rnn}
    \mathbf{h}_{t+1} = \mathbf{W}\mathbf{h}_t + \mathbf{U}\mathbf{x}_t + \mathbf{b}
\end{equation}

Through recursive expansion, the state at time $T$ can be written as:

\begin{equation} \label{eq:recursive_expansion}
    \mathbf{h}_T = \mathbf{W}^T\mathbf{h}_0 + \sum_{k=0}^{T-1} \mathbf{W}^k(\mathbf{U}\mathbf{x}_{T-1-k} + \mathbf{b})
\end{equation}

State saturation in linear RNNs occurs when:

\begin{enumerate}
    \item \textbf{Eigenvalue Domination}: If $\|\lambda_{\text{max}}(\mathbf{W})\| > 1$, where $\lambda_{\text{max}}$ denotes the eigenvalue with largest magnitude, the term $\mathbf{W}^T\mathbf{h}_0$ grows exponentially, causing state explosion.

    \item \textbf{Fixed Point Convergence}: If $\|\lambda_{\text{max}}(\mathbf{W})\| < 1$, the state converges to a fixed point:

    \begin{equation} \label{eq:fixed_point}
        \mathbf{h}_{\infty} = (\mathbf{I} - \mathbf{W})^{-1}(\mathbf{U}\bar{\mathbf{x}} + \mathbf{b})
    \end{equation}

    \noindent where $\bar{\mathbf{x}}$ represents the average input. This convergence limits the network's ability to capture new information.
\end{enumerate}

\subsection{Nonlinear RNNs}

For nonlinear RNNs with bounded activation functions (e.g., tanh, sigmoid), state saturation manifests differently. Consider the tanh activation:

\begin{equation} \label{eq:nonlinear_rnn}
    \mathbf{h}_{t+1} = \tanh(\mathbf{W}\mathbf{h}_t + \mathbf{U}\mathbf{x}_t + \mathbf{b})
\end{equation}

Saturation occurs through two primary mechanisms:

\begin{enumerate}
    \item \textbf{Activation Saturation}: When $\|\mathbf{W}\mathbf{h}_t + \mathbf{U}\mathbf{x}_t + \mathbf{b}\| \gg 1$, the tanh function saturates:

    \begin{equation} \label{eq:tanh_limits}
        \lim_{z \to \infty} \tanh(z) = 1, \quad \lim_{z \to -\infty} \tanh(z) = -1
    \end{equation}

    The gradient in saturated regions approaches zero:

    \begin{equation} \label{eq:tanh_gradient}
        \frac{\partial \tanh(z)}{\partial z} = 1 - \tanh^2(z) \approx 0
    \end{equation}

    \item \textbf{Dynamic Attractor Formation}: The nonlinear system forms attractors in state space described by the fixed-point equation:

    \begin{equation} \label{eq:attractor}
        \mathbf{h}^* = \tanh(\mathbf{W}\mathbf{h}^* + \mathbf{U}\bar{\mathbf{x}} + \mathbf{b})
    \end{equation}

    Once the state approaches these attractors, the network's capacity to encode new information diminishes.
\end{enumerate}

\subsection{Information Theoretic Perspective}

State saturation can be quantified through the mutual information between inputs and states:

\begin{equation} \label{eq:mutual_info}
    I(\mathbf{X}_{t-\tau:t}; \mathbf{h}_t) = H(\mathbf{h}_t) - H(\mathbf{h}_t|\mathbf{X}_{t-\tau:t})
\end{equation}

\noindent where $H$ is the entropy and  $\mathbf{X}_{t-\tau:t}$ represents the input sequence from time $t-\tau$ to $t$. As saturation occurs:

\begin{equation} \label{eq:info_limit}
    \lim_{t \to \infty} I(\mathbf{x}_t; \mathbf{h}_t|\mathbf{X}_{t-\tau:t-1}) \to 0
\end{equation}

\noindent indicating diminishing capacity to encode new information in the saturated state.





\subsection{Implications for Continuous Processing}
The mathematical analysis shows that while reset mechanisms successfully prevent state saturation, they introduce fundamental limitations in both information preservation and gradient propagation. These limitations become particularly problematic in continuous processing scenarios, where:
\begin{enumerate}
    \item The optimal reset timing cannot be determined a priori.
    \item Important temporal dependencies may span across reset points.
    \item The discontinuity in state evolution may introduce artifacts in the output sequence.
\end{enumerate}
\section{Training}

We implement all experiments using the AdamW optimizer with a batch size of 512 and an initial learning rate of 3e-3 across all experimental conditions. Hidden states were initialized using random sampling during reset operations. For the SNNs, we employ a Gaussian-like surrogate gradient function for backpropagation through the spiking activation function.

\end{appendices}

\end{document}